\documentclass[final]{cvpr}

\usepackage{times}
\usepackage{epsfig}
\usepackage{graphicx}
\usepackage{amsmath}
\usepackage{amssymb} 
  
\usepackage{tabularx}
\usepackage{booktabs}
\usepackage{multirow} 
\usepackage[accsupp]{axessibility}
\usepackage{enumitem} 
\usepackage[dvipsnames, table, x11names]{xcolor}
\usepackage[accsupp]{axessibility}  

\definecolor{citecolor}{HTML}{0071bc}
\usepackage[pagebackref=false,breaklinks=true,letterpaper=true,colorlinks,citecolor=citecolor,bookmarks=false]{hyperref}

\def\cvprPaperID{2309} 
\def\confYear{CVPR 2022} 

\usepackage[english]{babel}
\newtheorem{theorem}{Theorem}

\newtheorem{proof}{Proof}

\makeatletter
\def\@fnsymbol#1{\ensuremath{\ifcase#1\or *\or \dagger\or \ddagger\or
   \mathsection\or \mathparagraph\or \|\or \dagger\dagger
   \or \ddagger\ddagger \else\@ctrerr\fi}}
\makeatother

\begin{document}
\definecolor{init}{RGB}{153, 153, 153}
\definecolor{so}{RGB}{107, 142, 35}
\title{Knowledge Distillation as Efficient Pre-training: Faster Convergence, \\Higher Data-efficiency, and Better Transferability}


\author{
Ruifei He$^{1}$\thanks{Part of the work is done during an internship at ByteDance AI Lab.} , \ Shuyang Sun$^{2}$\thanks{Equal contribution.} , \ Jihan Yang$^{1\dagger}$, \ Song Bai$^{3}$\thanks{Corresponding author.} , \ Xiaojuan Qi$^{1\ddagger}$\\
$^{1}$The University of Hong Kong \ $^{2}$University of Oxford  \ $^{3}$ByteDance \\
{\tt\small \{ruifeihe, jhyang, xjqi\}@eee.hku.hk, kevinsun@robots.ox.ac.uk, songbai.site@gmail.com}
} 

\maketitle

\begin{abstract}
    Large-scale pre-training has been proven to be crucial for various computer vision tasks.
    However, with the increase of pre-training data amount, model architecture amount, and the private/inaccessible data, it is not very efficient or possible to pre-train all the model architectures on large-scale datasets. 
   In this work, we investigate an alternative strategy for pre-training, namely Knowledge Distillation as Efficient Pre-training (\textbf{KDEP}), aiming to efficiently transfer the learned feature representation from existing pre-trained models to new student models for future downstream tasks. 
   We observe that existing Knowledge Distillation (KD) methods are unsuitable towards pre-training since they normally distill the logits that are going to be discarded when transferred to downstream tasks. 
   To resolve this problem, we propose a feature-based KD method with non-parametric feature dimension aligning.
      Notably, our method performs comparably with supervised pre-training counterparts in 3 downstream tasks and 9 downstream datasets requiring \textbf{10$\times$} less data and \textbf{5$\times$} less pre-training time. Code is available at \url{https://github.com/CVMI-Lab/KDEP}.
  \end{abstract}

\vspace{-0.4cm}
\section{Introduction} \label{Sec:intro}
With the booming of large-scale datasets \cite{deng2009imagenet, mahajan2018exploring, OpenImages, sun2017revisiting, ridnik2021imagenet}, many computer vision tasks have benefitted significantly from pre-training in the past decade. In fact, it has been a de facto strategy to first pre-train on datasets like ImageNet \cite{deng2009imagenet} and then fine-tune on downstream tasks \cite{zhao2017pyramid, chen2017deeplab, he2017mask, yang2018learning, redmon2016you}, especially when the data of downstream tasks is scarce. 

However, the increasing pre-training data scale and the inaccessibility of private data~\cite{sun2017revisiting} render pre-training all architectures on large datasets not efficient or possible.
As well-trained deep neural networks are essentially condensed memory bank of datasets \cite{arpit2017closer, feldman2020neural}, we wonder whether the condensed data knowledge encoded into a pre-trained model can be leveraged to efficiently pre-train new architectures with only a relatively small set of pre-training data?

In this work, we propose Knowledge Distillation as Efficient Pre-training (KDEP), transferring the feature extraction capability of the teacher obtained from large-scale data, to the student model for solving future downstream tasks.
Note that KDEP is quite different from traditional Knowledge Distillation (KD) that only targets at distilling the knowledge of a given specific task to a student model.

\begin{figure} 
    \begin{center}
    \includegraphics[width=0.95\linewidth]{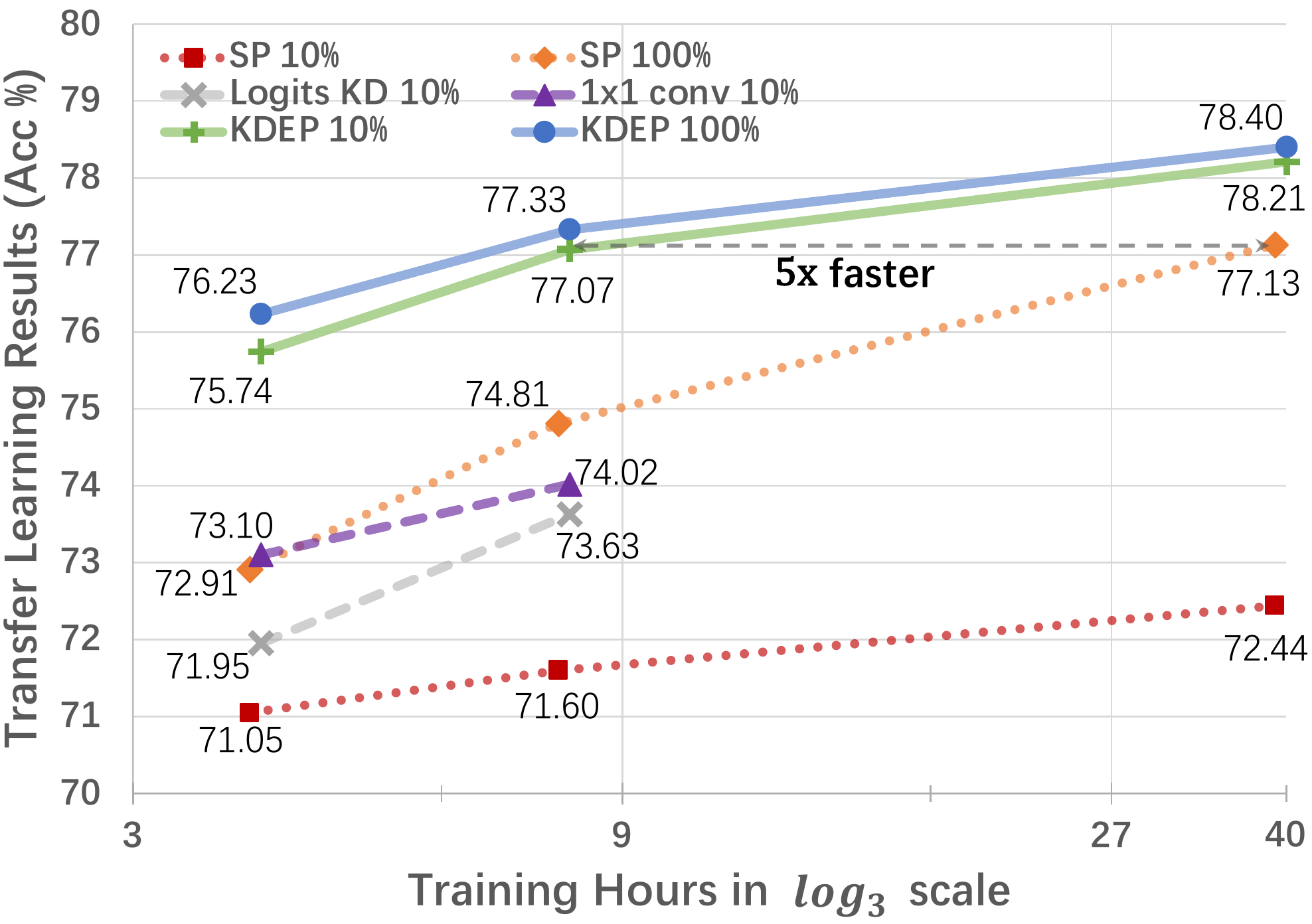} 
    \end{center}
    \vspace{-0.4cm}
       \caption{Transfer performance (averaged top-1 accuracy of four image classification tasks (details in Sec. \ref{Sec: exp})) compared to Supervised Pre-training (\textbf{SP}), traditional KD method (\textbf{logits KD}, \textbf{1$\times$1 conv}), and \textbf{KDEP} (SVD+PTS) with different data amount (10\% or 100\% ImageNet-1K data) and training schedules. }
    \label{fig: performance curve} 
    \vspace{-0.4cm} 
    \end{figure}

    \begin{figure}  
        \begin{center}
        \includegraphics[width=1.0\linewidth]{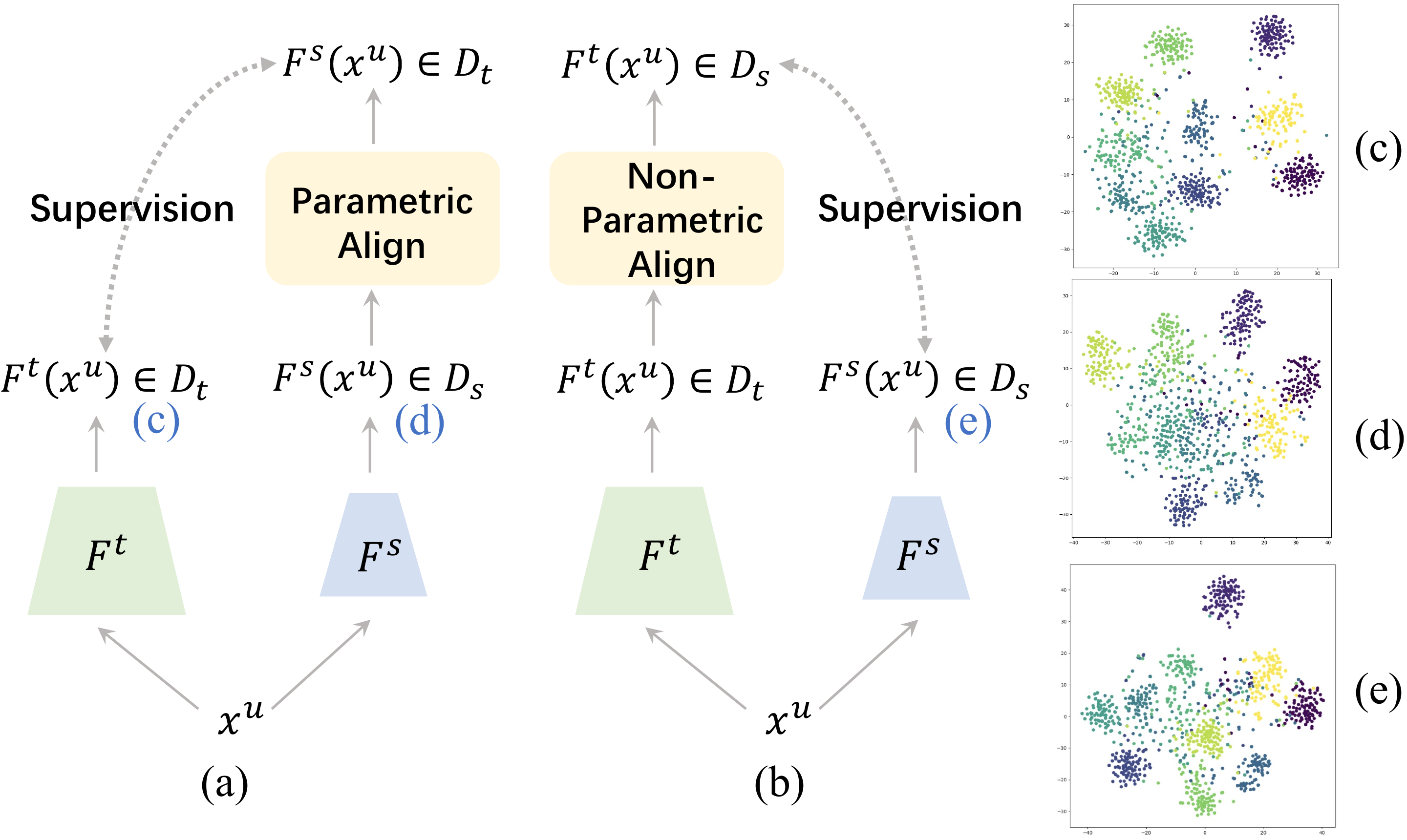} 
        \end{center}
        \vspace{-0.4cm}
           \caption{\textbf{Framework and visualization.} Here we illustrate the KDEP framework of feature-based KD with \textbf{(a)}:  
            parametric
           aligning; and \textbf{(b)}: non-parametric aligning. Notations refer to Sec. \ref{Sec: overview}. Notice that only the learned $F^s$ is to transfer to downstream tasks. Moreover, we visualize the feature representation of \textbf{(c)}: original teacher, \textbf{(d)}: student distilled by (a) with 1$\times$1 conv aligning, and \textbf{(e)}: student distilled by (b) with our SVD+PTS aligning. For visualization, we randomly sample 10 classes in ImageNet-1K and use 100 samples from each class. T-SNE \cite{van2008visualizing} is used for dimensionality reduction.} 
        \label{fig: framework vis} 
        \vspace{-0.5cm} 
        \end{figure} 

\noindent\textbf{Studies of existing KD methods for KDEP.} 
Our empirical studies show that existing KD methods such as logits KD \cite{hinton2015distilling} (\ie distilling the task-specific output logits) and feature-level KD \cite{heo2019comprehensive} lead to inferior performance (see Figure~\ref{fig: performance curve}: ``logits KD'' and ``1$\times$1 conv''), indicating that existing KD methods tailored to different tasks might be unable to fully leverage the knowledge condensed in the teacher model when pre-training new models with limited data and computation budget.

After further investigation, we conclude a potential issue of ``indirect feature learning'', where the supervision of distillation is not \emph{directly applied} on the feature extractor that will be transferred, but on a new learnable module added after it and jointly optimized with it, which turns the feature representation learning into an indirect process.  Specifically, ``logits KD'' applies the supervision on the classifier's output, where the learnable parameters of the classifier is jointly learned with the feature extractor. Although feature-level KD applies supervisions on the features, most feature-level KD methods adopt a parametric module to align the feature dimensions of teacher and student, 
typically a 1$\times$1 convolutional layer (``1$\times$1 conv'') \cite{romero2014fitnets,heo2019comprehensive,heo2019knowledge, ahn2019variational},
which again adds learnable parameters after the feature extractor and forms an indirect process. As shown in Figure \ref{fig: framework vis}d, we visualize the learned feature representation of feature-based KD with 1$\times$1 conv alignment: the learned feature representation fails to follow that of the teacher's (Figure \ref{fig: framework vis}c).
Consequently, the pre-trained models only deliver suboptimal transfer learning performance.

\noindent\textbf{Our Method.}
Motivated by the identified potential problem, our KDEP investigates non-parametric methods for aligning the feature dimensions to avoid indirect feature learning. 
Empirically, we found Singular Value Decomposition (SVD) works effectively by compressing\footnote{We focus on the setting that teachers having larger dimensions in our study, since this is more frequently met in real-world applications. Foundation models tend to have larger feature dimensions where models for deployment usually have smaller dimensions. } the features with  minimal information loss.

However, features processed by SVD will trigger the component domination effect \cite{chen2017makeup}, \ie the feature variances among channels are of great magnitude differences, and largely differ from those of normal DCNNs. This interferes the optimization of the network.
To further boost feature learning, we design a Power Temperature Scaling (PTS) function to reduce the variance differences while preserving the original relative magnitude, which tailors features from SVD for DCNNs. 
As illustrated in Figure \ref{fig: framework vis}e, with our SVD+PTS aligning method, the distilled student obtains feature representation similar to the original teacher's (Figure \ref{fig: framework vis}c) while matching the feature dimensions.
Notably, our method adds no learnable parameters, does not rely on the task loss or the logits loss \cite{hinton2015distilling}, and only use supervision from the teacher's penultimate feature (after global average pooling), which is more general for feature representation learning and allows more potential pre-trained teachers.

\noindent\textbf{Results.} 
Our major findings of KDEP are summarized in Figure \ref{fig: performance curve}: 
1) \textbf{Faster convergence}. Our method achieves comparable or better transfer learning results with only 10\% or 20\% training time than supervised pre-training (SP) on the whole ImageNet-1K dataset.
2) \textbf{Higher data-efficiency}. With only 10\% of ImageNet-1K unlabeled data (discard the labels) and an available pre-trained teacher model, our distilled student obtains better transfer learning results than SP with 100\% ImageNet-1K data. 
3) \textbf{Better Transferability}. Given the same computation budget and data amount as SP, our method achieves higher transfer learning performance.
With the proposed KDEP method, we could realize pre-training once and distilling it to all: distilling a pre-trained teacher (either to utilize an available pre-trained model or pre-train a certain architecture) to efficiently pre-train all other student models. 

\section{Related Works}

\noindent\textbf{Transfer Learning (TL)}, usually by fine-tuning a pre-trained model to a downstream task with labeled data, has become a common practice in machine learning problems and applications. To better understand TL, it can be separated into two steps: pre-training and transfer. 

Recent years have witnessed increasing successful works on pre-training, including supervised pre-training \cite{joulin2016learning, li2017learning, mahajan2018exploring, sun2017revisiting, kolesnikov2020big}, self-supervised pre-training \cite{chen2020simple, he2020momentum, caron2020unsupervised, grill2020bootstrap, chen2021exploring, zbontar2021barlow, ye2019unsupervised}, and semi-supervised pre-training \cite{xie2020self, pham2021meta, chen2020big}. 
Given the pre-trained models, the next step is to transfer the learned representation to a target task. Except from the widely used fine-tuning method \cite{yosinski2014transferable, agrawal2014analyzing}, there are also other methods proposed for better exploiting the knowledge absorbed in pre-training, such as L2-SP \cite{xuhong2018explicit}, DELTA \cite{li2019delta}, BSS \cite{chen2019catastrophic}, and Co-Tuning \cite{you2020co}.

While large-scale pre-training yield better representation and downstream performance,
the cost of pre-training is also rapidly increasing \cite{sun2017revisiting, dosovitskiy2020image, kolesnikov2020big}.
Therefore, we hope to propose an orthogonal strategy for pre-training, distilling a pre-trained model to pre-train different student models.

\noindent\textbf{Knoledge Distillation (KD)} has been developed as an effective way for model compression and acceleration, and various methods mainly falls into three research streams: response-based \cite{bucilu2006model,hinton2015distilling, zhou2021rethinking}, feature-based \cite{heo2019knowledge, romero2014fitnets, komodakis2017paying, huang2017like, passalis2018learning,kim2018paraphrasing, jin2019knowledge,chen2021cross, heo2019comprehensive}, and relation-based \cite{yim2017gift, lee2018self, passalis2020heterogeneous, liu2019knowledge, park2019relational, chen2020learning, tung2019similarity, peng2019correlation, tian2019contrastive} methods. 

Response-based methods usually take the final outputs called logits as supervision and use a temperature factor to adjust the smoothness \cite{hinton2015distilling}. 
Though the logits introduce ``dark knowledge'' into training and show improved results, response-based KD leaves out large information of the intermediate features, which are found to be crucial for representation learning \cite{romero2014fitnets}.

Feature-based KD was first introduced by Fitnets \cite{romero2014fitnets}, using intermediate feature maps as hints to improve KD performance. Following Fitnets, attention maps \cite{komodakis2017paying}, neuron selectivity patterns \cite{huang2017like}, paraphraser \cite{kim2018paraphrasing}, route constraint \cite{jin2019knowledge}, and activation boundary \cite{heo2019knowledge} are also proposed to better utilize feature-level knowledge. Heo {\etal} \cite{heo2019comprehensive} investigate different design aspects of feature-based KD, and propose margin ReLU, Pre-ReLU distillation position, and a partial L2 loss function. Chen {\etal} \cite{chen2021cross} use an attention mechanism to adaptively assign proper teacher layers for each student layer.

Relation-based methods further boost the performance by utilizing the relationships between different layers or data samples. FSP \cite{yim2017gift} uses inner products between features of two layers as a flow of solution process, but it restricts to the same feature map sizes between teacher and student. Lee {\etal} \cite{lee2018self} also utilize the correlations between feature maps by using Radial Basis Function, and apply SVD on spatial dimensions to both teacher and student's feature maps to avoid the mismatch in spatial resolutions.  However, they still need the dimension of feature maps to be equal for teacher and student. In contrast, our method can adapt to teacher and student pairs with different feature map resolutions and dimensions. Another line in relation-based methods is utilizing the relation between data samples, where different mechanisms have been proposed, including instance relationship graph \cite{liu2019knowledge}, similarity matrix \cite{tung2019similarity}, similarity probability distribution \cite{passalis2020heterogeneous} and so on.

However, traditional KD methods only focus on a single task and transfer task-specific knowledge, while our KDEP focuses on the transferability of the distilled student, which distinguishes us from most previous KD methods.
To our best knowledge, only Li {\etal} \cite{li2021representation} try to utilize KD for student's transferability. They show traditional KD would hurt the transferability of the student, and propose a multi-head, multi-task distillation method using an unlabeled proxy dataset and a generalist teacher to improve downstream performance of the distilled student. 

Nevertheless, their method needs the student model initialized by ImageNet pre-trained weights, multiple teachers fine-tuned on the domains which are related to the downstream domains, and a multi-head, multi-task training procedure, which violates our efficient pre-training setting. On the contrary, our method optimizes the student from scratch, and only needs a single generalist teacher, an unlabeled dataset with a simple yet efficient training pipeline to achieve comparable transfer learning performance with supervised pre-training.

\section{KDEP}

\subsection{Overview} \label{Sec: overview}
 
We define the KDEP settings as below: given a teacher model $F^t$ (pre-trained on a large-scale dataset $\mathcal{D}$) and a set of unlabeled examples $ \mathcal{D}_u = \{x^u_i\}^{N_u}_{i=1}$ ($N_u$ is the number of unlabeled images), our goal is to pre-train a student model $F^s$ to generalize well on various downstream tasks. Note that the dataset scale of $ \mathcal{D}_u$ could be magnitude smaller than $\mathcal{D}$ and only $ \mathcal{D}_u$ is available during the student training. Since we focus on feature representation learning rather than tailoring the model to a specific task, both $F^t$ and $F^s$ yield the feature representations instead of task-specific logits. We denote the shape of $F^t(x^u_i)$, $F^s(x^u_i)$ as $D_t$, $D_s$. The training objective of the KDEP method is
\vspace{-0.25cm} 
\begin{equation}
    \frac{1}{N_u} \sum_{i=1}^{N_u}\mathcal{L}(F^t(x^u_i),F^s(x^u_i)), 
    \label{eq:1}
\vspace{-0.3cm} 
\end{equation}
where $\mathcal{L}$ is the $L_2$ loss.
To meet the demand of our proposed KDEP, several under-explored obstacles are needed to be addressed.

The first is a known issue for feature-based KD, that is the feature dimension mismatch (\ie, $D_t \neq D_s$) between the teacher and student. In our study, we found the frequently adopted strategy, to add a parametric module like a 1$\times$1 conv, is sub-optimal for our KDEP settings. 
Instead, we demonstrate that non-parametric methods (\eg SVD) are more effective than 1$\times$1 conv for aligning the dimensions.
Analysis and details will unfold in Sec.\ref{Sec: Aligning Feature Dimensions}.

The second is a byproduct of our non-parametric feature dimension aligning method: the feature statistics after the alignment would differ from normal DCNNs. Hence, we study several mechanisms for correcting the feature statistics and conclude them as a transformation module. We will further elaborate on the design choices in Sec. \ref{Sec: Transformation Module}.

The third is still an open issue even after our exploration: What is a good teacher for KDEP? Our empirical studies show that stronger models are not necessarily better teachers, and we find the compactness of the teacher's feature distribution to be a crucial indicator (see Sec. \ref{Sec: Compactness}). We hope we could inspire more future works on this topic.

\subsection{Aligning Feature Dimensions} \label{Sec: Aligning Feature Dimensions} 
Motivated by the indirect feature learning problem, we propose non-parametric feature dimension aligning methods with several variants. 
Concretely, previous parametric methods add a parametric module to align $D_s$ toward the supervision $D_t$
(ref. Figure \ref{fig: framework vis}a). In contrast, as presented in Figure \ref{fig: framework vis}b, 
we apply a non-parametric method to project $D_t$ to $D_s$, which can then serve as the supervision directly.

We investigate three variants for non-parametric aligning methods: channel selection, interpolation, and SVD. Along this line, SVD stands out thanks to its power of effectively compressing the feature-level knowledge and maintaining as much information as possible. Detailed experimental results would be included in Sec. \ref{Sec: exp}.

\noindent\textbf{Pre-ReLU distillation feature position} has been used in previous feature-based KD \cite{heo2019knowledge, heo2019comprehensive} and shown improved results. In our methods, we distill the features before the ReLU activation function for
one more consideration, that SVD's outputs contain both negative and positive values.

\begin{figure} 
    \begin{center}
    \includegraphics[width=0.85\linewidth]{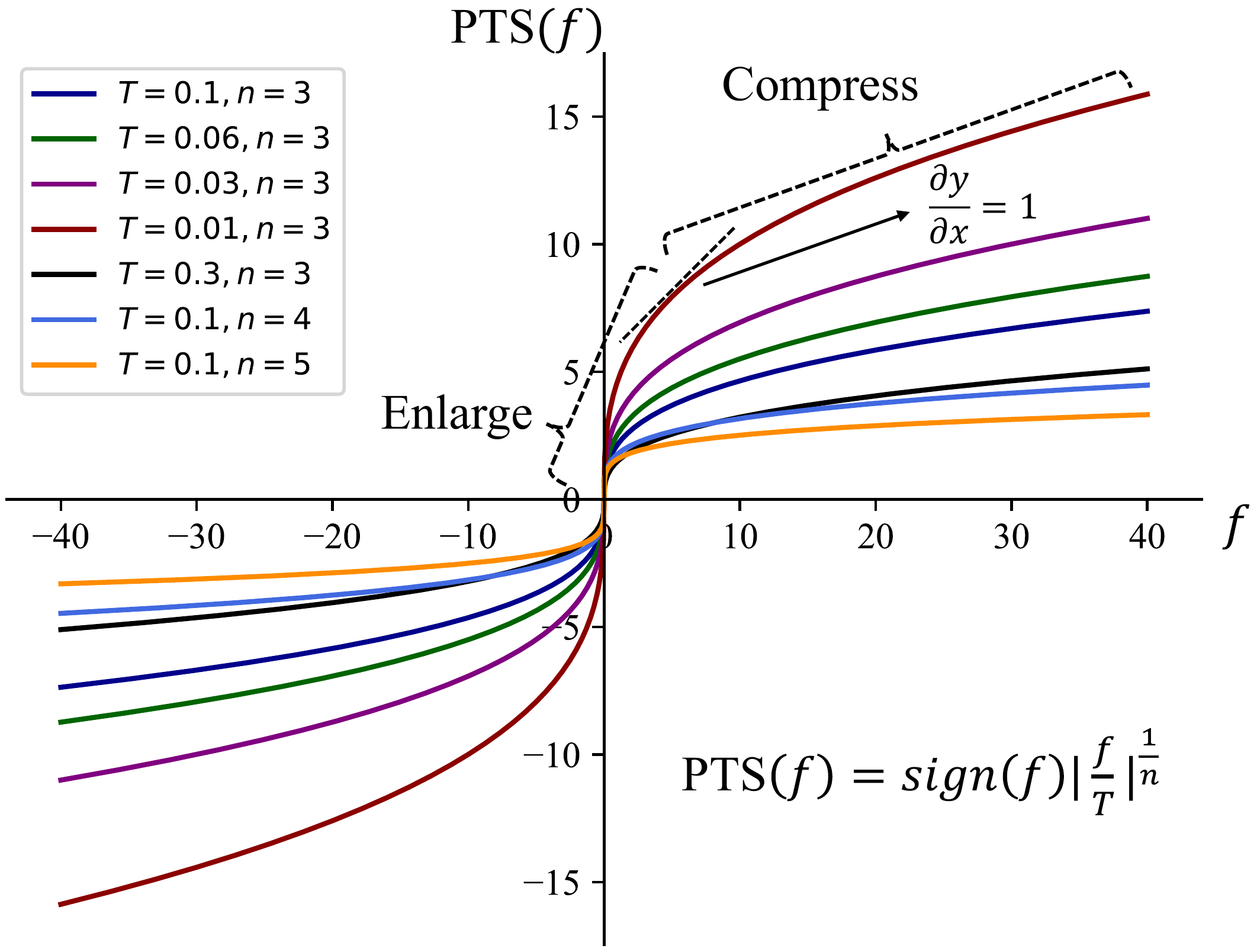} 
    \end{center}
    \vspace{-0.3cm}
       \caption{We propose PTS function for reducing Std Ratio while preserving the original relative magnitude. Curves of different values of T and n are shown.}
    \label{fig: pts func} 
    \vspace{-0.4cm} 
    \end{figure}

\subsection{Transformation Module} \label{Sec: Transformation Module}
While SVD effectively compresses the features with unnoticeable information loss, it brings along difficulties for optimization. 
After the SVD alignment, the feature variances of different channels are of magnitude differences, whereas those of normal DCNNs are usually within the same magnitude. Concretely, we define the term Std Ratio as the largest standard deviation (Std) to the smallest Std among all feature channels at the penultimate features across all training data samples.
According to our study, we found the Std Ratio of features after SVD is usually over 10$\times$ larger than that of normal DCNNs.

As a result, the $L_2$ loss tends to be dominated by the feature channels with the largest variances, leaving the minor ones under-fitted, for which we provide theoretical analysis below. We view the value of each feature channel from the teacher as a random variable ($T$), which is of zero mean after SVD. Also, since the student is optimized from random initialization \cite{he2015delving}, we regard each feature channel of the student also as a random variable ($S$) with zero mean. 
As shown in Theorem \ref{theorm} and Proof \ref{proof}, we state and prove that the mathematical expectation of the $L_2$ loss from each feature channel increases monotonically with the Std of the teacher's feature channel, which explains the difficulty of learning from teacher with large Std Ratio.

\begin{theorem} \label{theorm}
    Given two independent random variables with normal distribution \(T \sim N(0,\sigma^2)\) and \(S \sim N(0,\sigma_s^2)\),  then \(F(\sigma)=\mathbb{E}[(T-S)^2]\) is monotonically increasing (\(\sigma>0\)).
\end{theorem}

\vspace{-0.3cm}
\begin{proof} (Detailed Proof in Appendix.) \label{proof}
    \begin{align*}\begin{split}
    & F(\sigma)= \int_{ - \infty }^{ + \infty } \int_{ - \infty }^{ + \infty } {(t-s)^2 \cdot P(t,s)\mathrm{d}t\mathrm{d}s} \\
    &= \frac{1}{\sqrt{2\pi}\sigma} \int_{ - \infty }^{ + \infty } {e^{\frac{-t^2}{2\sigma^2}} (t^2+\sigma_s^2)\mathrm{d}t} = \sigma^2+\sigma_s^2  \\
    &\frac{\mathrm{d}F(\sigma)}{\mathrm{d}\sigma} = 2\sigma > 0 \Rightarrow \text{monotonically increasing}
     \end{split} \end{align*}
\end{proof}
\vspace{-0.01cm}

Therefore, we propose to reduce the Std Ratio after SVD to normal ranges of DCNNs by a transformation module. A simple method of Scale Normalization (SN) has been used in previous works \cite{chen2017makeup, lee2018self}, which divides each feature channel by its corresponding Std to ensure each channel has similar scales. Similarly, we also experiment with a variant of SN: rather than dividing the corresponding Std to obtain similar scales, we scale each channel's Std to match the top-$D_s$ Std of features before SVD (target Std), which we name as Std Matching (SM).
However, both SN and SM are local transformations that transform channel-wisely, and thus may fail to preserve the original relative magnitude between different feature channels (example in Appendix). 

To match statistics without hurting the original relative magnitude, we propose to use a global non-decreasing transformation function that can reduce the Std Ratio while maintaining the relative magnitude. Concretely, we control the value ranges by a temperature parameter $T$ similar to logits KD \cite{hinton2015distilling}, and then apply a power operation while preserving the signs. We refer to the function as Power Temperature Scaling (PTS), which is as follows:
\vspace{-0.3cm}
\begin{equation}
    \text{PTS}(f) = sign(f) \lvert  \frac{f}{T}  \lvert^{\frac{1}{n}}, 
    \label{eq:PTS}
    \vspace{-0.3cm}
\end{equation}
where n is the parameter of the exponent and $f$ is the input value.
As shown in Figure \ref{fig: pts func}, the PTS function could enlarge small values and compress large values, while globally non-decreasing, and thus fulfills our goal of matching normal statistics and preserving the relative magnitude.

\subsection{Teacher Selection} \label{Sec: Compactness}

In this section, we further explore how to select a good teacher for KDEP. Naturally, we would consider utilizing stronger models as the teacher, where we study and compare several paradigms of potential stronger models:
\vspace{-0.25cm}
\begin{itemize}   
    \setlength{\itemsep}{0pt}  
    \setlength{\parsep}{0pt} 
    \setlength{\parskip}{0pt}
    \item \textbf{Standard SP}: the most frequent supervised pre-training strategy that pre-trains a model on ImageNet-1K; we use the architecture of ResNet50 (R50) \cite{he2016deep} as teacher for this study;
    \item SP with more data: we experiment with an available Microsoft (\textbf{MS}) Vision R50\footnote{https://pypi.org/project/microsoftvision/} pre-trained with four datasets (over 40 million data): ImageNet-22K, COCO, and two webly-supervised datasets \cite{yao2016extracting}. The provided weights only contain the feature extractor; 
    \item Pre-training with unlabeled data: we explore a semi-weakly supervised pre-trained R50 (\textbf{SWSL});
    \item Distilled models: we use a \textbf{MEAL V2} R50 trained by distillation on ImageNet-1K.
    \item Advanced architecture: we experiment with a state-of-the-art architecture Swin Transformer \cite{liu2021swin} and use \textbf{Swin-B} pre-trained with ImageNet-22K and fine-tuned on ImageNet-1K.
\end{itemize}  
\vspace{-0.35cm}

We empirically found that stronger models (\ie higher performance on ImageNet-1K benchmark) do not necessarily achieve better KDEP performance (\ie distilled student's transferability under the KDEP setting), which resonates with previous findings in KD \cite{muller2019does, menon2021statistical} that more accurate teachers may distill worse. To investigate the reasons, we visualize the feature representation of different teachers (ref. Figure \ref{fig: compact}), and surprisingly found that the KDEP performance has a strong correlation with the teacher's feature compactness (compactness means the feature distribution of data samples of the same class lying tightly in the feature space while those of different classes far from each other). Detailed analysis and results are in Sec. \ref{sec: ablation teachers}.

\section{Experiments} \label{Sec: exp}

\subsection{Experimental Setup}
\vspace{-0.1cm}
\paragraph{Datasets and downstream tasks.} For the proposed KDEP settings, we use the ImageNet-1K \cite{deng2009imagenet} dataset as unlabeled data by abandoning the labels, and we use 10\% or 100\% of the dataset for different settings. 

To evaluate the transfer learning performance of the models, we evaluate on three popular downstream tasks: image classification, semantic segmentation, and object detection. For image classification, we select four diverse datasets to study the transferability: CIFAR-100 \cite{krizhevsky2009learning}, CUB-200-2011 \cite{wah2011caltech}, DTD \cite{cimpoi2014describing}, and Caltech-256 \cite{griffin2007caltech}. For semantic segmentation, we use three widely used datasets: Cityscapes \cite{cordts2016cityscapes}, PASCAL VOC 2012 (VOC12) \cite{everingham2015pascal}, and ADE20K \cite{zhou2019semantic}. For object detection, we evaluate the transfer performance on two benchmarks: PASCAL VOC \cite{everingham2010pascal}, and COCO \cite{lin2014microsoft}.  

\vspace{-0.6cm}
\paragraph{Teacher-Student (T-S) pair.} We experiment with two different teacher-student pairs,  R50 $\rightarrow$ ResNet18 (R18), and R50 $\rightarrow$ MobileNetV2 (MNV2) \cite{sandler2018mobilenetv2}, representing knowledge transfer between similar and dissimilar networks, respectively. For the teacher model, we conduct our main experiments with standard SP R50 and MS R50. 

\vspace{-0.5cm}
\paragraph{Comparison methods.} We mainly compare KDEP with supervised pre-training (SP). We denote the student supervised pre-trained with all ImageNet-1K data for 90 epochs as SP oracle (\textbf{SP. o.}), and with fewer data or for shorter schedule as SP baseline (\textbf{SP. b.}) in each setting.

\vspace{-0.5cm}
\paragraph{Implementation details.} We implement our method using PyTorch \cite{paszke2019pytorch} and all experiments are conducted using four 32G V100 GPUs.
For studying KDEP, we explore different settings with various data amounts and training schedules. For the 10\% ImageNet-1K data setting, we set the training epochs to 90 or 180; when using 100\% ImageNet-1K data, we train for 9 or 18 epochs to verify fast convergence, and for 90 epochs to further boost performance, where 90 epochs with all ImageNet-1K data is the standard supervised pre-training schedule \cite{salman2020adversarially}.  
For all downstream tasks, we use the same schedule and evaluation protocols for all models for a fair comparison. 
More elaborated implementation details are given in the Appendix.

\vspace{-0.5cm}
\paragraph{Evaluation.} 
We report the top-1 accuracy, mean Intersection over Union (mIoU), and AP, AP$_\text{50}$, AP$_\text{75}$ for classification, segmentation, and detection, respectively.
All results are averaged over at least three trials. Time refers to the pre-training time of SP or KDEP on four 32G V100 GPUs.
\subsection{Main Results} \label{Sec: main results}
In this section, we compare our best KDEP method (SVD+PTS) with supervised pre-training under different data amounts and training schedules. We evaluate the transferability on all 9 transfer learning tasks covering image classification, semantic segmentation and object detection. With our extensive experimental results (\eg Table \ref{tab: main result R18}), we demonstrate that Knowledge Distillation can be used as an effective way of pre-training, outperforming standard supervised pre-training with fewer training data and shorter training schedules. In the following, we explore the transferability, data-efficiency, and convergence speed of KDEP respectively under different setups.

Note that all KDEP methods use the MS R50 as the teacher in this section. Results with more teachers are in our ablation studies. Also, due to the length limit, we show the results of R18 as student in our paper, and MNV2 as student in the Appendix, where similar results are achieved.

\noindent\textbf{Exploring transferability under 10\% data with short schedules.}
In this setting, we use only 10\% ImageNet-1K data, which is a total of 128k images randomly sampled from the original 1.28 million images. We pre-train for 90 epochs or 180 epochs with KDEP or SP. As shown in Table \ref{tab: main result R18}, KDEP significantly outperforms its SP counterparts in different settings. 
 
We take the transfer performance in classification as an example to illustrate and analyze in this setting and the settings following unless noted. 
SP models' performance drops significantly when pre-trained with only 10\% data (77.13$\rightarrow$71.05). Even increasing the training schedule to 180 epochs (71.05$\rightarrow$71.60) or 900 epochs (71.05$\rightarrow$72.44) fails to eliminate such performance drop.
In contrast, with only 90 epochs, KDEP (75.74) largely bridges the gap between supervised pre-training baselines and oracle. Increasing the schedule to 180 epochs further closes the gap to an unnoticeable drop (77.07 \vs 77.13) while only using 10\% data and around 20\% training time. Similar results are also observed in segmentation and detection results as well as the R50 $\rightarrow$ MNV2 pair (in Appendix).
 
\noindent\textbf{Exploring transferability under the same data amount and schedule as standard SP.}
We further explore the KDEP performance given the same data amount and training schedules as standard SP. In Table \ref{tab: main result R18}, with 100\% data and 90 training epochs, KDEP produces models with much stronger transferability (78.40) than standard SP (77.13), while only adding minuscule computation costs.
 
\begin{table}[htbp] 
    \centering
    \vspace{-0.2cm}
    \begin{footnotesize}
        \setlength\tabcolsep{2.5pt}
        \begin{tabular}{ccccccccc}
            \bottomrule[1pt]
            \multirow{2}{*}{Method}  & \multirow{2}{*}{Data} & \multirow{2}{*}{Epoch} & \multirow{2}{*}{\shortstack{Time\\(/h)}} & \multicolumn{5}{c}{\textbf{Classification (Acc \%)}}   \\ 
             && && Caltech &  DTD & CUB & CIFAR & Avg\\
            \hline
            \textcolor{init}{rand. init.} & \textcolor{init}{-} & \textcolor{init}{-} & \textcolor{init}{-}& \textcolor{init}{55.27} & \textcolor{init}{45.16} & \textcolor{init}{55.89} & \textcolor{init}{77.34} & \textcolor{init}{58.42}  \\
            \hline
            SP. b. &10\% & 90 & 3.9 & 68.83 & 66.17 & 69.93 & 79.29 & 71.05 \\ 
            \rowcolor{Gray!16}
            KDEP &10\%& 90 & 4.0 & 75.33 & 71.80 & 74.20 & 81.61 & 75.74 \\
            SP. b. &100\% & 9 & 3.9 & 71.27 & 68.14 & 71.97 & 80.27 & 72.91 \\ 
            \rowcolor{Gray!16}
            KDEP &100\%& 9 & 4.0 & 75.42 & 72.15 & 75.11 & 82.22 & 76.23 \\
            \hline
            SP. b. &10\%& 180 & 7.8 & 70.09 & 66.14& 70.84 & 79.34 & 71.60 \\ 
            \rowcolor{Gray!16}
            KDEP &10\%& 180 & 8.0 & 77.15 & 72.67 & 75.99 & 82.23 & \textbf{77.07} \\
            SP. b. &100\%& 18 & 7.8 & 74.01 & 69.18 & 74.63 & 81.43 & 74.81 \\
            \rowcolor{Gray!16}
            KDEP &100\%& 18 & 8.0 & 77.29 & 73.07 & 76.50 & 82.47 & \textbf{77.33} \\
            \hline 
            SP. b. &10\%& 900 & 39 & 71.10 & 67.02 & 72.16 & 79.49 & 72.44 \\
            \rowcolor{Gray!16}
            KDEP &10\%& 900 & 40 &  79.00 & 74.28 & 76.89 & 82.64 & \textbf{78.21} \\
            \textcolor{so}{SP. o.} &\textcolor{so}{100\%}& \textcolor{so}{90} & \textcolor{so}{39} & \textcolor{so}{77.18} & \textcolor{so}{71.81} & \textcolor{so}{77.44} & \textcolor{so}{82.08} & \textcolor{so}{77.13}\\
            \rowcolor{Gray!16}
            KDEP &100\%& 90 & 40 & 79.08 & 74.34 & 77.29 & 82.89 & \textbf{78.40}  \\
            \toprule[0.8pt]
        \end{tabular}
    \end{footnotesize}

    \begin{footnotesize}
        \setlength\tabcolsep{2.5pt}
        \begin{tabular}{cccccccc}
            \bottomrule[1pt]
            \multirow{2}{*}{Method}  & \multirow{2}{*}{Data} & \multirow{2}{*}{Epoch} & \multirow{2}{*}{\shortstack{Time\\(/h)}} & \multicolumn{4}{c}{\textbf{Segmentation (mIoU \%)}}   \\ 
             && && Cityscapes & VOC12 & ADE20K  & Avg\\
            \hline
            \textcolor{init}{rand. init.} & \textcolor{init}{-} & \textcolor{init}{-} & \textcolor{init}{-}& \textcolor{init}{57.87} & \textcolor{init}{49.46} & \textcolor{init}{31.37} & \textcolor{init}{46.23}  \\
            \hline
            SP. b. &10\% & 90 & 3.9 & 69.56 & 68.30 &  34.39 & 57.41  \\ 
            \rowcolor{Gray!16}
            KDEP &10\% & 90 & 4.0 & 70.41 &72.34 & 36.09 & 59.61 \\
            SP. b. &100\% & 9 & 3.9 & 67.92 & 70.05 & 35.27 & 57.75\\
            \rowcolor{Gray!16}
            KDEP &100\%& 9 & 4.0 & 69.73 & 72.43 & 36.17 & 59.44 \\
            \hline
            SP. b. &10\% & 180 & 7.8 & 69.89 & 69.79 & 35.03 & 58.24 \\ 
            \rowcolor{Gray!16}
            KDEP &10\% & 180 & 8.0 & 70.27 & 72.82 & 36.60 & \textbf{59.90} \\
            SP. b. &100\% & 18 & 7.8 & 70.19 & 71.02 & 35.19 & 58.80 \\
            \rowcolor{Gray!16}
            KDEP &100\%& 18 & 8.0 & 70.90 & 73.82 & 36.73 & \textbf{60.48}\\
            \hline 
            SP. b. &10\% & 900 & 39& 69.85 & 69.55 & 35.52 & 58.31 \\
            \rowcolor{Gray!16}
            KDEP &10\%& 900 & 40 & 71.93 & 74.28 & 37.30 & \textbf{61.17}\\
            \textcolor{so}{SP. o.} &\textcolor{so}{100\%} & \textcolor{so}{90} & \textcolor{so}{39}& \textcolor{so}{71.01} & \textcolor{so}{73.13}& \textcolor{so}{36.02} & \textcolor{so}{60.05}\\
            \rowcolor{Gray!16}
            KDEP &100\%& 90 & 40 & 71.39 & 73.75 & 36.97& \textbf{60.70} \\
            \toprule[0.8pt]
        \end{tabular}
    \end{footnotesize}

    \begin{footnotesize}
        \setlength\tabcolsep{2.6pt}
        \begin{tabular}{cccccccccc}
            \bottomrule[1pt]
            \multirow{3}{*}{Method}  & \multirow{3}{*}{Data} & \multirow{3}{*}{Epoch} & \multirow{3}{*}{\shortstack{Time\\(/h)}} & \multicolumn{6}{c}{\textbf{Detection}}   \\ 
             &&& & \multicolumn{3}{c}{VOC0712} & \multicolumn{3}{c}{COCO}\\
             && && AP& AP$_\text{50}$ & AP$_\text{75}$ & AP& AP$_\text{50}$ & AP$_\text{75}$ \\
            \hline
            \textcolor{init}{rand. init.} & \textcolor{init}{-} & \textcolor{init}{-} & \textcolor{init}{-}& \textcolor{init}{26.7} & \textcolor{init}{52.5} & \textcolor{init}{23.1} & \textcolor{init}{25.2} & \textcolor{init}{41.9} & \textcolor{init}{26.3}  \\
            \hline
            SP. b. &10\%& 90 & 3.9 & 39.8&	69.7&	39.1 & 27.6	&45.3	&28.8 \\ 
            \rowcolor{Gray!16}
            KDEP &10\%& 90 & 4.0 & \textbf{41.9}	&72.4&	41.7 & 28.6	&46.5 &	29.9 \\
            SP. b. &100\% & 9 & 3.9 &  40.4&	70.5&	40.2 & 27.5	&45.3&	28.6 \\
            \rowcolor{Gray!16}
            KDEP &100\%& 9 & 4.0 & \textbf{42.5}&	73.3	&43.3&28.8&	46.9&	30.1  \\
            \hline
            SP. b. &10\%& 180 & 7.8 & 39.4&	69.6&	38.7 & 28.2	&45.9&	29.4 \\ 
            \rowcolor{Gray!16}
            KDEP &10\%& 180 & 8.0 & \textbf{43.4}&	73.8&	43.8&  \textbf{29.2}	&47.4&	30.6 \\
            SP. b. &100\% & 18 & 7.8 & 41.2&	71.5&	40.7 & 28.1	&46.1	&29.3 \\
            \rowcolor{Gray!16}
            KDEP &100\%& 18 & 8.0 & \textbf{43.3}	&73.6&	44.2 & \textbf{29.3}	& 47.5 &	30.8 \\
            \hline 
            SP. b. &10\% & 900 & 39& 39.3&	69.9&	38.8& 28.5	&47.0&	29.9\\
            \rowcolor{Gray!16}
            KDEP &10\%& 900 & 40 &\textbf{42.8}	&73.5	&43.4& \textbf{29.9}&	48.4	&31.7 \\
            \textcolor{so}{SP. o.} &\textcolor{so}{100\%}& \textcolor{so}{90} & \textcolor{so}{39}& \textcolor{so}{41.8}	& \textcolor{so}{72.6}&	\textcolor{so}{41.6} & \textcolor{so}{29.0}	& \textcolor{so}{47.3}	&\textcolor{so}{30.4} \\
            \rowcolor{Gray!16}
            KDEP &100\%& 90 & 40 & \textbf{42.8}	&73.9&	43.4 & \textbf{29.7}&	48.2&	31.3  \\
            
            \toprule[0.8pt]
        \end{tabular}
    \end{footnotesize}

    \vspace{-0.01cm}  
    \caption{\textbf{KDEP \vs SP, R50 $\rightarrow$ R18, fine-tuned on various tasks.} KDEP refers to our SVD+PTS method. Note that COCO is used for the teacher's pre-training while not for the student's.}
    \vspace{-0.6cm}
    \label{tab: main result R18}
\end{table}

\noindent\textbf{Exploring data-efficiency.}
Here, we again only use 10\% data but extend the training schedule to 900 epochs, which aligns the training iterations with the standard supervised pre-training setting.
In Table \ref{tab: main result R18}, KDEP could consistently outperform the supervised oracle on all 9 tasks with only 10\% ImageNet-1K data, 
which verifies our initial idea that models pre-trained on large-scale data could transfer the condensed data knowledge to other architectures even without using full pre-training data.

\noindent\textbf{Exploring convergence speed.} 
For further exploring convergence speed, we compare KDEP and SP under different data amounts (\ie 10\% and 100\% ImageNet-1K data) as the training time increases. As shown in Figure \ref{fig: performance curve}, the transfer performance of SP increases almost uniformly with the training time, while KDEP shows a favourable characteristic of fast convergence under both data amounts. Notably, KDEP produces comparable or superior transfer performance as standard SP with 5$\times$ fewer training time.

\begin{figure*}
    \begin{center}
    \includegraphics[width=0.9\linewidth]{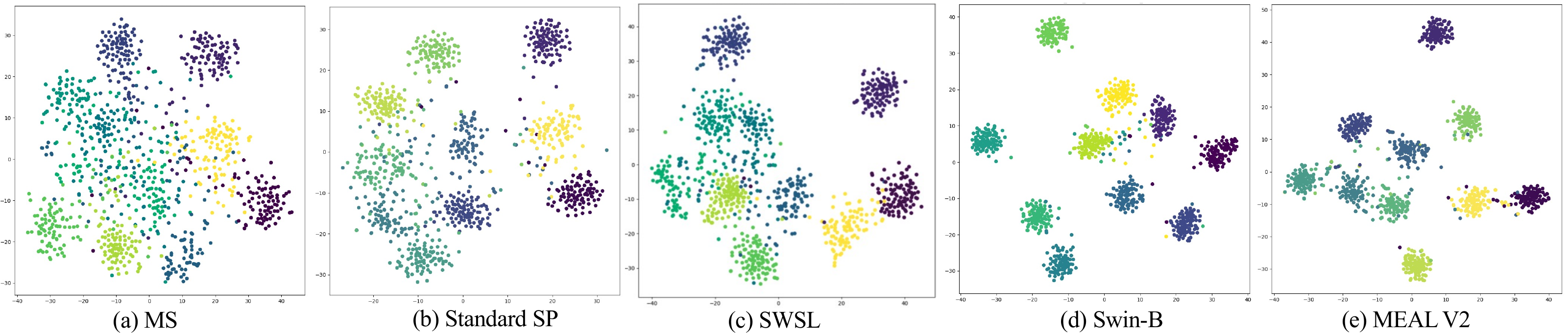} 
    \end{center}
    \vspace{-0.4cm}
       \caption{Visualization (same as in Figure \ref{fig: framework vis}) of feature distributions. Order ranked by compactness: diverse (left) to compact (right).}
    \label{fig: compact}
    \vspace{-0.3cm}
    \end{figure*} 

\subsection{Ablation Studies}
For our KDEP settings, we notice a strong correlation between the transfer learning results of different tasks in Sec. \ref{Sec: main results}. Hence, we use four image classification tasks to evaluate the transferability in our ablation studies. All results in our ablation studies are the averaged top-1 accuracy over four classification tasks.

\noindent\textbf{Ablation study on feature dimension aligning methods.}
In this ablation study, we aim to investigate the effectiveness of various feature dimension aligning methods for KDEP.
Firstly, we compare parametric methods with non-parametric methods under the 10\% data and short schedule of 90 epochs setting with three Teacher-Student (T-S) pairs. 
For parametric methods, we experiment with three 1$\times$1 conv variants: Post-ReLU, Pre-ReLU and the one in \cite{heo2019comprehensive} (details in Appendix). 
For non-parametric methods, we explore channel selection (CS.var, CS.rand), interpolation, and SVD (details in Appendix).
As shown in Table \ref{tab: ablation align}, among three variants of 1$\times$1 conv methods, the two that adopts Pre-ReLU feature distillation position give better performance. Interestingly, non-parametric methods consistently outperform all variants of the 1$\times$1 conv method with significant gains. Moreover, SVD produces the best performance under various T-S pairs.

Further, we study the data-efficiency and convergence speed of parametric methods under various KDEP settings, with results shown in Table \ref{tab: ablation 1x1 conv}. For data-efficiency, parametric methods suffer even trained for 900 epochs with 10\% data, showing low data-efficiency. For settings verifying convergence speed (\ie 100\% data with 9 or 18 epochs), parametric methods perform similarly with the supervised pre-training baselines. The lacking of high data-efficiency and fast convergence characteristics 
may be caused by the added learnable module that hinders feature learning.

\begin{table}[htbp] 
    \centering
    \vspace{-0.2cm}
    \begin{small}
        \setlength\tabcolsep{2pt} 
        \begin{tabular}{cccc}
            \bottomrule[1pt]
            Teacher & Standard SP R50 & MS R50  & MS R50    \\ 
            Student & R18 & R18 & MNV2 \\
            \hline
            1$\times$1 conv (Post-ReLU) &  71.37 & 72.43 & 71.12 \\
            1$\times$1 conv (Pre-ReLU) & 71.01 & 73.10 &73.72 \\
            1$\times$1 conv (\cite{heo2019comprehensive})& 71.35 & 72.98& 73.62  \\
            \hline
            CS.var & 73.54 & 74.60 &74.92\\
            CS.rand & 73.47 & 74.37 &74.90 \\
            Interpolation & 73.27& 74.81 & 74.96 \\
            SVD & \textbf{74.29}& \textbf{75.09} & \textbf{75.07}\\
            \toprule[0.8pt] 
        \end{tabular}
    \end{small}
    \vspace{-0.01cm}  
    \caption{\textbf{Ablation study on feature dimension aligning methods.} Setting: 10\% data and 90 epochs. CS: ``Channel Selection".}
    \vspace{-0.3cm}
    \label{tab: ablation align}
\end{table}

\begin{table}[htbp] 
    \centering
    \vspace{-0.3cm}
    \begin{footnotesize}
        \setlength\tabcolsep{3.5pt}
        \begin{tabular}{ccccc}
            \bottomrule[1pt]
            Method & Data & Epoch & \shortstack{Time (/h)} & Acc (\%) \\
            \hline
            SP. b. &10\% & 90 & 3.9  & 71.05 \\ 
            KDEP (1$\times$1 conv, Pre-ReLU) &10\%& 90 & 4.0 & 73.10  \\
            KDEP (1$\times$1 conv, \cite{heo2019comprehensive}) &10\%& 90 & 4.0 & 72.98  \\
            \rowcolor{Gray!16}
            KDEP (SVD, PTS) &10\%& 90 & 4.0 &  \textbf{75.74} \\
            \hline
            SP. b. &100\% & 9 & 3.9 &  72.91 \\ 
            KDEP (1$\times$1 conv, Pre-ReLU) &100\%& 9 & 4.0 & 73.24 \\
            KDEP (1$\times$1 conv, \cite{heo2019comprehensive}) &100\%& 9 & 4.0 & 73.15 \\
            \rowcolor{Gray!16}
            KDEP (SVD+PTS) &100\%& 9 & 4.0 & \textbf{76.23} \\
            \hline
            SP. b. &10\%& 180 & 7.8 &  71.60 \\ 
            KDEP (1$\times$1 conv, Pre-ReLU) &10\%& 180 & 8.0 &  74.02 \\
            KDEP (1$\times$1 conv, \cite{heo2019comprehensive}) &10\%& 180 & 8.0 &  73.90 \\
            \rowcolor{Gray!16}
            KDEP (SVD, PTS) &10\%& 180 & 8.0 &  \textbf{77.07} \\
            \hline
            SP. b. &100\%& 18 & 7.8  & 74.81 \\
            KDEP (1$\times$1 conv, Pre-ReLU) &100\%& 18 & 8.0  & 74.32 \\
            KDEP (1$\times$1 conv, \cite{heo2019comprehensive}) &100\%& 18 & 8.0  & 73.78 \\
            \rowcolor{Gray!16}
            KDEP (SVD+PTS) &100\%& 18 & 8.0  & \textbf{77.33} \\
            \hline 
            SP. b. &10\%& 900 & 39 &  72.44 \\
            KDEP (1$\times$1 conv, Pre-ReLU) &10\%& 900 & 40  & 75.29 \\
            KDEP (1$\times$1 conv, \cite{heo2019comprehensive}) &10\%& 900 & 40   & 75.33 \\
            \rowcolor{Gray!16}
            KDEP (SVD+PTS) &10\%& 900 & 40   & \textbf{78.21} \\ 
            \hline 
            \textcolor{so}{SP. o.} &\textcolor{so}{100\%}& \textcolor{so}{90} & \textcolor{so}{39} &  \textcolor{so}{77.13}\\
            \toprule[0.8pt]
        \end{tabular}
    \end{footnotesize}
    \vspace{-0.01cm}  
    \caption{\textbf{Ablation study on the 1$\times$1 conv method.} T-S pair: MS R50 $\rightarrow$ R18.}
    \vspace{-0.2cm}
    \label{tab: ablation 1x1 conv}
\end{table}

\noindent\textbf{Ablation study on transformation module.}
We explore the effectiveness of three mechanisms of the transformation module as introduced in Sec. \ref{Sec: Transformation Module}. Concretely, we experiment with two T-S pairs in the setting of 10\% data and short schedules of 90 or 180 epochs. As shown in Table \ref{tab: ablation transformation}, PTS works as a competitive method across different setups, while scale normalization (SN) and Std Matching (SM) also achieve performance gains upon the SVD method, which shows the importance of matching statistics while preserving original relative magnitude.

Moreover, we conduct hyper-parameter analysis to study the sensitiveness of the hyper-parameters in the PTS function. From Table \ref{tab: para PTS}, both $T$ and $n$ have a relatively wide range that could bring performance gains upon SVD (``w.o'' in the Table), which illustrates the robustness of the PTS function.

\begin{table}[htbp] 
    \centering
    \vspace{-0.1cm}
    \begin{small}
        \setlength\tabcolsep{10pt} 
        \begin{tabular}{ccccc}
            \bottomrule[1pt]
            T $\rightarrow$ S & \multicolumn{2}{c}{R50 $\rightarrow$ R18}   & \multicolumn{2}{c}{R50 $\rightarrow$ MNV2}   \\ 
            Epoch& 90 & 180 & 90 & 180 \\

            \hline
            SVD &  75.09 & 76.28& 75.07 & 76.28 \\
            SVD+SN & 75.34 & 76.45 & \textbf{75.43} & 76.33\\
            SVD+SM& \textbf{75.87} & \underline{76.89} & 75.21 & \underline{76.35} \\
            SVD+PTS& \underline{75.74} & \textbf{77.07} & \underline{75.37} & \textbf{76.53}\\
            
            \toprule[0.8pt] 
        \end{tabular}
    \end{small}
    \vspace{-0.05cm}  
    \caption{\textbf{Ablation study on the transformation module.} Setting: 10\% data with 90 or 180 epochs. R50: MS. SN: Scale Normalization. SM: Std Matching. Bold: best. Underlined: second-best.}
    \vspace{-0.1cm}
    \label{tab: ablation transformation}
\end{table}

\begin{table}[htbp] 
    \centering
    \vspace{-0.4cm}

    \begin{footnotesize}
        \setlength\tabcolsep{3pt} 
        \begin{tabular}{cccccccccc}
            \bottomrule[1pt]
            $T$ & 0.01 & 0.03 & 0.06 & 0.1 & 0.3 & 0.5 & 0.7 & 0.9 & w.o. \\
            \hline
            Acc & 75.12 & 75.36 & 75.53 & \textbf{75.74} & 75.51 & 75.21 & 75.29 & 75.07 & 75.09 \\
            \toprule[0.8pt] 
        \end{tabular}
    \end{footnotesize}

    \begin{footnotesize}
        \setlength\tabcolsep{8pt} 
        \begin{tabular}{ccccccc}
            \bottomrule[1pt]
            $n$ & 2 & 3 & 4 & 5 & 6 & w.o. \\
            \hline
            Acc & 75.06 & \textbf{75.74} & 75.69 & 75.61 & 75.19 & 75.09 \\
            \toprule[0.8pt] 
        \end{tabular}
    \end{footnotesize}

    \vspace{-0.01cm}  
    \caption{\textbf{Hyper-parameter analysis on the PTS function.} Setting: 10\% data with 90 epochs. T-S pair: MS R50 $\rightarrow$ R18. When varying $T$, we fix $n=3$. When varying $n$, we fix $T=0.1$.}
    \vspace{-0.3cm}
    \label{tab: para PTS}
\end{table}

\noindent\textbf{Ablation Studies with different teacher models.} \label{sec: ablation teachers}
As shown in Table \ref{tab: ablation teachers}, teachers with higher accuracy on ImageNet benchmark do not necessarily lead to better KDEP performance. Intriguingly, combining the visualization results in Figure \ref{fig: compact}, we notice a strong correlation between the teacher's feature compactness and KDEP performance, that compact feature representation suffers to serve as a good teacher, whereas diverse ones produce superior results.  

\begin{table}[htbp] 
    \centering
    \vspace{-0.2cm}
    \begin{small}
        \setlength\tabcolsep{2.5pt} 
        \begin{tabular}{cccccc}
            \bottomrule[1pt]
            Teacher & Standard SP & MS & SWSL & Swin-B & MEAL V2  \\ 
            ImageNet & 75.77 & 73.85 & 81.12 & \textbf{84.81} &  80.68 \\
            \hline
            SVD& 74.29 & \textbf{75.09} & 73.75 & 72.053 & 70.33 \\
            SVD+PTS& 74.79 & \textbf{75.74} & 74.27 & 72.252 & 72.41  \\
            \toprule[0.8pt] 
        \end{tabular}
    \end{small}
    \vspace{-0.01cm}  
    \caption{\textbf{Ablation study on different teacher models.} Setting: 10\% data with 90 epochs. ImageNet: ImageNet val set top-1 Acc.}
    \vspace{-0.2cm}
    \label{tab: ablation teachers}
\end{table}

\noindent\textbf{Ablation study on multiple layer feature-KD.}
Our method only distills the penultimate layer feature to transfer knowledge, whereas multiple layer feature-based KD has also been introduced in the literature. Here, we study the effectiveness of multiple layer feature-based KD for our KDEP settings. From Table \ref{tab: ablation multi}, we observe that multiple layer feature-based KD may be more beneficial when teacher and student are of similar architectures (\ie R50 $\rightarrow$ R18), but may cause performance degradation for dissimilar architecture pairs (\ie R50 $\rightarrow$ MNV2) due to potential semantic mismatch of intermediate features. 

\begin{table}[htbp] 
    \centering
    \vspace{-0.0cm}
    \begin{small}
        \setlength\tabcolsep{1.3pt} 
        \begin{tabular}{cllll}
            \bottomrule[1pt]
            T $\rightarrow$ S  & \multicolumn{2}{l}{MS R50 $\rightarrow$ R18}   & \multicolumn{2}{l}{MS R50 $\rightarrow$ MNV2}   \\ 
            Epoch& 90 & 180 & 90 & 180 \\
            \hline
            SVD+PTS& 75.74 & 77.07 & 75.37 & 76.53\\
            SVD+PTS+ML& 76.11 \scriptsize{\textcolor{teal}{(+0.37)}}  & 77.21 \scriptsize{\textcolor{teal}{(+0.14)}} & 75.20 \scriptsize{\textcolor{red}{(-0.17)}} & 76.25 \scriptsize{\textcolor{red}{(-0.28)}} \\
            \toprule[0.8pt] 
        \end{tabular}
    \end{small}
    \vspace{-0.01cm}  
    \caption{\textbf{Ablation study on multiple layer feature-based KD.} Setting:  10\% data with 90 or 180 epochs. ML: multiple layer feature-based KD.}
    \vspace{-0.3cm}
    \label{tab: ablation multi} 
\end{table}
 
\noindent\textbf{Ablation study on logits KD.} 
We also experiment with logits KD for our KDEP settings. Since the best MS R50 teacher does not contain weights to produce logits, we study with the second-best standard SP R50 teacher. As shown in Table \ref{tab: ablation logit}, logits KD leads to inferior performance compared with our feature-based KD under various data amounts and training schedules. More importantly, logits KD may fail to utilize potential better teachers trained on more data, which in our case could produce much better KDEP results (see ``Ours (SVD+PTS)$^\dagger$'' in Table \ref{tab: ablation logit}). 

\begin{table}[htbp] 
    \centering
    \vspace{-0.1cm}
    \begin{small}
        \setlength\tabcolsep{5pt}
        \begin{tabular}{ccccc}
            \bottomrule[1pt] 
            Method & Data & Epoch & \shortstack{Time (/h)} & Acc (\%) \\
            \hline
            SP. b. &10\% & 90 & 3.9  & 71.05 \\ 
            Logits &10\%& 90 & 4.0 & 71.95  \\
            \rowcolor{Gray!16}
            Ours (SVD+PTS) &10\%& 90 & 4.0 & 74.79 \\
            \rowcolor{Gray!16}
            Ours (SVD+PTS)$^\dagger$  &10\%& 90 & 4.0 & \textbf{75.74}  \\
            \hline
            SP. b. &10\%& 180 & 7.8 &  71.60 \\ 
            Logits  &10\%& 180 & 8.0 &  73.63 \\
            \rowcolor{Gray!16}
            Ours (SVD+PTS) &10\%& 180 & 8.0 &  76.02 \\
            \rowcolor{Gray!16}
            Ours (SVD+PTS)$^\dagger$ &10\%& 180 & 8.0 &  \textbf{77.07} \\
            \hline 
            \textcolor{so}{SP. o.} &\textcolor{so}{100\%}& \textcolor{so}{90} & \textcolor{so}{39} &  \textcolor{so}{77.13}\\
            Logits  &100\%& 90 & 40 &  77.19 \\
            \rowcolor{Gray!16} 
            Ours (SVD+PTS) &100\%& 90 & 40 &  77.66 \\
            \rowcolor{Gray!16}
            Ours (SVD+PTS)$^\dagger$ &100\%& 90 & 40 &  \textbf{78.40} \\
            \toprule[0.8pt] 
        \end{tabular}
    \end{small}
    \caption{\textbf{Ablation study on logits KD.} T-S pair: Standard SP R50 $\rightarrow$ R18. $^\dagger$: MS R50 (provided weights do not contain classifiers', so we cannot apply logits KD with this model).}
    \vspace{-0.3cm}
    \label{tab: ablation logit}
\end{table}

\subsection{Discussion} \label{Sec: discuss}
\noindent\textbf{Varying KDEP data $\mathcal{D}_u$.}
In the above studies, we use 10\% or 100\% ImageNet-1K data as $ \mathcal{D}_u$, which is known to have large diversity. We further experiment with different $\mathcal{D}_u$: (1) object-level data. We use the four downstream image classification datasets respectively; (2) scene-level data. We use COCO and ADE20K respectively. Since these different $\mathcal{D}_u$ are of different sizes, we keep the training iterations same as the 10\% data with 90 epochs setting. Results are shown in Table \ref{tab: ablation D_u}, where we conclude that the KDEP performance could relate to different aspects of $\mathcal{D}_u$: data amount, data diversity, and image context (object or scene level). 

Harvesting the largest data amount and diversity, ImageNet-1K yields the best KDEP performance with object-centric samples.
Meanwhile, COCO holds similar data scale and fairly diverse scene-level images, producing the second-best results.
On the contrary, constrained by the fewest images of only texture patterns (low diversity), DTD leads to the worst results.
We suggest that enlarging data amount and diversity is beneficial for KDEP, and object-level images are more favourable for our current KDEP method.
However, we believe scene-level images could be better utilized by further leveraging the characteristic of scene context, which we leave as future work.
\begin{table}[htbp] 
    \centering
    \vspace{-0.1cm}
    \begin{small}
        \setlength\tabcolsep{2.5pt} 
        \begin{tabular}{ccccccc}
            \bottomrule[1pt]
            $\mathcal{D}_u$ & $N_u$ & Caltech &  DTD & CUB & CIFAR & Avg\\
            \hline
            Caltech &15k & 72.77 & 67.14 & 67.48 & 80.03 & 71.86 \\ 
            DTD &3.8k &54.32 & 64.97 & 62.43 & 74.79 & 64.13  \\
            CUB &6.0k &62.74 & 61.72 &78.82 &  78.31 & 70.40 \\
            CIFAR &50k& 61.83 & 56.36 & 61.19 & 81.48 & 65.21 \\
            \hline
            COCO &118k & 72.34 & 69.55 & 71.68 & 79.87 & 73.36 \\ 
            ADE20K &20k &67.66 & 67.38 & 68.20 & 78.90 & 70.54 \\
            \hline 
            10\% ImageNet-1K &128k& 75.33 & 71.80 & 74.20 & 81.61 & \textbf{75.74} \\

            \toprule[0.8pt] 
        \end{tabular}
    \end{small}
    \vspace{-0.01cm}  
    \caption{\textbf{Varying $\mathcal{D}_u$.} Setting: the same training iterations as 10\% ImageNet-1K data with 90 epochs. T-S pair: MS R50 $\rightarrow$ R18.}
    \vspace{-0.2cm}
    \label{tab: ablation D_u}
\end{table}
  
\noindent\textbf{Computation cost.} In most of our experimental results, we provide the training time of KDEP and supervised pre-training, where only unnoticeable extra training time is added. Moreover, the GPU memory usage during KDEP is also similar to supervised pre-training since we do not require gradients for the teacher. Yet, our teacher model is R50, and additional computation costs may increase when using larger teachers that inquires more inference time and GPU memory. Still, the pre-training time could be largely reduced compared with supervised pre-training.  
 
\section{Conclusion}
We have present KDEP, an orthogonal strategy for pre-training new models.   
With extensive experimental results, our simple yet efficient feature-based KD method has shown promising results for KDEP, offering several favourable characteristics: Faster Convergence, Higher Data-efficiency and Better Transferability. Without bells and whistles, KDEP achieves comparable transfer performance as supervised pre-training with only 10x less data and 5x less training time. 

\vspace{-0.4cm}  
\paragraph{Limitations and Broader Impact.}
The KDEP performance largely relys on a suitable teacher model, where how to obtain such a teacher still needs further investigation. In our study, we found the compactness of feature distribution could be an important indicator, from which we would hope the community could release models not only with compact features (usually tailored to ImageNet task) but also with diverse feature distributions (pre-trained with large-scale diversified data). Moreover, we hope our work could inspire more research and applications on KDEP, which we believe is very useful for academic research and practical usage. 

\vspace{-0.3cm}
\section*{Acknowledgement}
\vspace{-0.1cm}
This work has been supported in part by Hong Kong Research Grant Council - Early Career Scheme (Grant No. 27209621), HKU Startup Fund, HKU Seed Fund for Basic Research, and SmartMore donation fund.

{\small
\bibliographystyle{ieee_fullname}
\bibliography{egbib}
}

\clearpage
\appendix

\definecolor{init}{RGB}{153, 153, 153}
\definecolor{so}{RGB}{107, 142, 35}

\renewcommand\thefigure{\thesection.\arabic{figure}}
\renewcommand\thetable{\thesection.\arabic{table}}
\setcounter{figure}{0} 
\setcounter{table}{0} 

\setcounter{table}{0}
\renewcommand{\thetable}{S.\arabic{table}}

\centerline{\large{\textbf{Outline}}}
\vspace{+0.3cm}
In this supplementary file, we first provide more results in Sec. \ref{sec: more results}: results of MobileNetV2 as student in Sec. \ref{sec: mnv2}, detailed proof of Theorem \textcolor{red}{1} in Sec. \ref{sec: proof}, comparison of previous feature-based KD methods in Sec. \ref{sec: compare}, and an example of local transformations breaking the original relative magnitude in Sec. \ref{sec: eg}. Further, we offer the elaborated implementation details for the KDEP setups and downstream task setups in Sec. \ref{sec: imple}.

 
\section{More Results} \label{sec: more results}

\subsection{Main Results: MobileNetV2 as Student.} \label{sec: mnv2}
Due to the length limit of the main paper, we show the results of MobileNetV2 as student in Table \ref{taba: mnv2}. Similar results have been achieved with MobileNetV2 (MNV2) as student compared to R18 as student, which shows the generalization of the proposed KDEP method.

\begin{table}[htbp] 
    \centering
    \vspace{-0.01cm}
    \begin{footnotesize}
        \setlength\tabcolsep{2.5pt} 
        \begin{tabular}{ccccccccc}
            \bottomrule[1pt]
            \multirow{2}{*}{Method}  & \multirow{2}{*}{Data} & \multirow{2}{*}{Epoch} & \multirow{2}{*}{\shortstack{Time\\(/h)}} & \multicolumn{5}{c}{\textbf{Classification (Acc \%)}}   \\ 
             && && Caltech &  DTD & CUB & CIFAR & Avg\\
            \hline
            \textcolor{init}{rand. init.} & \textcolor{init}{-} & \textcolor{init}{-} & \textcolor{init}{-}& \textcolor{init}{51.77} & \textcolor{init}{57.34} & \textcolor{init}{60.44} & \textcolor{init}{76.66} & \textcolor{init}{61.55}  \\ 
            \hline
            SP. b. &10\% & 90 & 4.2 & 66.58 & 65.72 & 71.09 & 78.60 & 70.50 \\ 
            \rowcolor{Gray!16}
            KDEP &10\%& 90 & 4.3 & 74.34 & 71.84 & 74.24 & 81.06 & 75.37  \\
            SP. b. &100\% & 9 & 4.2 & 68.09 & 67.514 & 73.33 & 78.95 & 71.97 \\
            \rowcolor{Gray!16} 
            KDEP &100\%& 9 & 4.3 & 74.48 & 72.51 & 74.76 & 81.47 & 75.81 \\
            \hline
            SP. b. &10\%& 180 & 8.4 &  69.23 & 67.33 & 73.62 & 79.50 & 72.42\\ 
            \rowcolor{Gray!16}
            KDEP &10\%& 180 & 8.6 & 75.56 & 73.29 & 75.28 & 81.98 & \textbf{76.53} \\
            SP. b. &100\%& 18 & 8.4 & 71.83 & 69.60 & 74.64 & 80.13 & 74.05 \\
            \rowcolor{Gray!16}
            KDEP &100\%& 18 & 8.6 &  76.06 & 73.14 & 76.00 & 82.15 & \textbf{76.83}\\
            \hline 
            SP. b. &10\%& 900 & 42 & 69.93 & 67.59 & 72.74 & 79.83 & 72.52\\
            \rowcolor{Gray!16}
            KDEP &10\%& 900 & 43 &  77.66 & 73.05 & 76.06 & 82.53 & \textbf{77.32}\\
            \textcolor{so}{SP. o.} &\textcolor{so}{100\%}& \textcolor{so}{90} & \textcolor{so}{42} & \textcolor{so}{76.43} & \textcolor{so}{72.26} & \textcolor{so}{76.34} & \textcolor{so}{81.91} & \textcolor{so}{76.74} \\
            \rowcolor{Gray!16}
            KDEP &100\%& 90 & 43 &  78.57 & 73.94 & 76.40 & 82.73& \textbf{77.91} \\
            \toprule[0.8pt] 
        \end{tabular}
    \end{footnotesize}

    \begin{footnotesize}
        \setlength\tabcolsep{2.5pt}
        \begin{tabular}{cccccccc}
            \bottomrule[1pt]
            \multirow{2}{*}{Method}  & \multirow{2}{*}{Data} & \multirow{2}{*}{Epoch} & \multirow{2}{*}{\shortstack{Time\\(/h)}} & \multicolumn{4}{c}{\textbf{Segmentation (mIoU \%)}}   \\ 
             && && Cityscapes & VOC12 & ADE20K  & Avg\\
            \hline
            \textcolor{init}{rand. init.} & \textcolor{init}{-} & \textcolor{init}{-} & \textcolor{init}{-}& \textcolor{init}{40.33} & \textcolor{init}{39.23} & \textcolor{init}{23.07} & \textcolor{init}{34.21}   \\
            \hline
            SP. b. &10\% & 90 & 4.2 & 60.92 & 62.60 & 29.17 & 50.89  \\ 
            \rowcolor{Gray!16}
            KDEP &10\% & 90 & 4.3 & 63.50 & 67.28 & 31.46 & 54.08 \\
            SP. b. &100\% & 9 & 4.2 & 61.42 & 64.31 & 29.61 & 51.78 \\
            \rowcolor{Gray!16}
            KDEP &100\%& 9 & 4.3 & 63.53 & 68.17 & 32.00 & 54.57\\
            \hline
            SP. b. &10\% & 180 & 8.4 & 61.68 & 64.65 & 29.95 & 52.09 \\ 
            \rowcolor{Gray!16}
            KDEP &10\% & 180 & 8.6 & 64.32 & 68.73 & 31.92 & \textbf{54.99}  \\
            SP. b. &100\% & 18 & 8.4 & 62.23 & 65.82 & 29.55 & 52.53\\
            \rowcolor{Gray!16}
            KDEP &100\%& 18 & 8.6 &64.23 & 69.32 & 32.41& \textbf{55.32} \\
            \hline 
            SP. b. &10\%& 900 & 42 & 61.87 & 64.95 & 31.07 & 52.63\\
            \rowcolor{Gray!16}
            KDEP &10\%& 900 & 43 &  63.89 & 70.45 & 32.23 & \textbf{55.52}\\
            \textcolor{so}{SP. o.} &\textcolor{so}{100\%} & \textcolor{so}{90} & \textcolor{so}{42}& \textcolor{so}{64.16} & \textcolor{so}{69.48} & \textcolor{so}{31.69} & \textcolor{so}{55.11}\\
            \rowcolor{Gray!16}
            KDEP &100\%& 90 & 43 & 64.72 & 71.07 & 32.39 & \textbf{56.06} \\
            \toprule[0.8pt]
        \end{tabular}
    \end{footnotesize}

    \begin{footnotesize}
        \setlength\tabcolsep{2.5pt}
        \begin{tabular}{cccccccccc}
            \bottomrule[1pt]
            \multirow{3}{*}{Method}  & \multirow{3}{*}{Data} & \multirow{3}{*}{Epoch} & \multirow{3}{*}{\shortstack{Time\\(/h)}} & \multicolumn{6}{c}{\textbf{Detection} }  \\ 
             &&& & \multicolumn{3}{c}{VOC0712} & \multicolumn{3}{c}{COCO}\\
             && && AP& AP$_\text{50}$ & AP$_\text{75}$ & AP& AP$_\text{50}$ & AP$_\text{75}$ \\
            \hline
            \textcolor{init}{rand. init.} & \textcolor{init}{-} & \textcolor{init}{-} & \textcolor{init}{-}& \textcolor{init}{31.4} & \textcolor{init}{56.7} & \textcolor{init}{30.2} & \textcolor{init}{25.8}& \textcolor{init}{43.6} & \textcolor{init}{26.9}\\
            \hline
            SP. b. &10\%& 90 & 4.2 & 42.5&	71.3	&44.0&  27.7&	46.4&	29.0 \\
            \rowcolor{Gray!16} 
            KDEP &10\%& 90 & 4.3 & \textbf{46.7}&	75.7&	49.0& \textbf{29.7}&	48.8&	31.2  \\
            SP. b. &100\% & 9 & 4.2 & 43.0	&71.9&	44.0& 27.9&	46.4&	29.0  \\
            \rowcolor{Gray!16}
            KDEP &100\%& 9 & 4.3 & \textbf{47.1}	&76.3&	49.6& \textbf{30.2}&	49.8	&31.9  \\
            \hline
            SP. b. &10\%& 180 & 8.4 &43.0&	71.4&	45.0 &  27.6&	46.2&	29.0 \\
            \rowcolor{Gray!16} 
            KDEP &10\%& 180 & 8.6 & \textbf{47.0}	&75.9&	49.9& \textbf{30.0}&	49.4	&31.5 \\
            SP. b. &100\% & 18 &8.4 & 43.7&	72.6	&45.4& 27.9	&46.6	&29.3  \\
            \rowcolor{Gray!16}
            KDEP &100\%& 18 & 8.6 &  \textbf{47.2}	&76.3&	50.0& \textbf{30.4}	&50.1&	32.0\\
            \hline 
            SP. b. &10\%& 900 & 42 & 44.0	&73.1	&45.6 & 28.7&	47.9&	30.1 \\
            \rowcolor{Gray!16}
            KDEP &10\%& 900 & 43 &\textbf{47.0}	&76.0&	49.8&\textbf{29.7}	&49.2&	31.4  \\
            \textcolor{so}{SP. o.} &\textcolor{so}{100\%}& \textcolor{so}{90} & \textcolor{so}{42}& \textcolor{so}{45.5}	&\textcolor{so}{75.0}	&\textcolor{so}{47.6}& \textcolor{so}{29.6}	&\textcolor{so}{49.1}	&\textcolor{so}{31.3} \\
            \rowcolor{Gray!16}
            KDEP &100\%& 90 & 43 & \textbf{46.8}	&76.5	&49.4 & \textbf{30.0}&	49.6&	31.4  \\
            
            \toprule[0.8pt]
        \end{tabular}
    \end{footnotesize}

    \vspace{-0.01cm}  
    \caption{\textbf{KDEP \vs SP, R50 $\rightarrow$ MNV2, fine-tuned on various tasks.} KDEP refers to our SVD+PTS method.}
    \vspace{-0.1cm}
    \label{taba: mnv2}
\end{table}

\subsection{Detailed Proof of Theorem 1} \label{sec: proof}

\textit{Given two independent random variables with normal distribution \(T \sim N(0,\sigma^2)\) and \(S \sim N(0,\sigma_s^2)\),  then \(F(\sigma)=\mathbb{E}[(T-S)^2]\) is monotonically increasing (\(\sigma>0\)).}


\begin{proof} (Detailed version) 
    \begin{align*}\begin{split}
    & F(\sigma)=\mathbb{E}[(T-S)^2] \\
    & = \int_{ - \infty }^{ + \infty } \int_{ - \infty }^{ + \infty } {(t-s)^2 \cdot P(t,s)\mathrm{d}t\mathrm{d}s} \\
    &= \int_{ - \infty }^{ + \infty } \int_{ - \infty }^{ + \infty } {(t-s)^2 \cdot P(t) \cdot P(s)\mathrm{d}t\mathrm{d}s} \\
    &= \int_{ - \infty }^{ + \infty } \int_{ - \infty }^{ + \infty } {(t-s)^2 \cdot  \frac{1}{2\pi\sigma\sigma_s}e^{\frac{-s^2}{2\sigma_s^2}}e^{\frac{-t^2}{2\sigma^2}}\mathrm{d}t\mathrm{d}s} \\
    &= \frac{1}{\sqrt{2\pi}\sigma} \int_{ - \infty }^{ + \infty } e^{\frac{-t^2}{2\sigma^2}} (\int_{ - \infty }^{ + \infty } {\frac{1}{\sqrt{2\pi}\sigma_s} (t-s)^2 \cdot e^{\frac{-s^2}{2\sigma_s^2}}\mathrm{d}s)\mathrm{d}t} \\
    &= \frac{1}{\sqrt{2\pi}\sigma} \int_{ - \infty }^{ + \infty } {e^{\frac{-t^2}{2\sigma^2}} (t^2+\sigma_s^2)\mathrm{d}t} = \sigma^2+\sigma_s^2 \\
    &\frac{\mathrm{d}F(\sigma)}{\mathrm{d}\sigma} = 2\sigma > 0 \Rightarrow \text{monotonically increasing}
     \end{split} \end{align*}
\end{proof}

\subsection{Compare other feature-based KD methods.} \label{sec: compare}
Here, we show the KDEP results of some traditional feature-based KD methods
that are developed for distilling knowledge to improve student's performance for a specific task instead of its transferability. 
Also, these methods all require task label loss which violates our setting of an unlabeled dataset.
Hence, we don't include them in our paper to compare for fairness.
As shown in Table \ref{tab: fkd}, previous feature-based KD methods largely rely on logit KD loss and task label loss, and perform inferiorly with only feature-level clues.
\begin{table}[htbp] 
    \centering
    \begin{small}
        \setlength\tabcolsep{0.4pt} 
        \begin{tabular}{cccccccc}
            \bottomrule[1pt]
            Method & SP & FitNet \cite{romero2014fitnets} & AT \cite{komodakis2017paying} & NST \cite{huang2017like} & AB \cite{heo2019knowledge} & Heo \cite{heo2019comprehensive} & Ours   \\ 
            \hline
            Acc & 71.05 & 72.43 & 67.84 & 67.05 & 71.66 & 72.98 & 75.74 \\
            \toprule[0.8pt] 
        \end{tabular}
    \end{small}
    \vspace{-0.2cm}  
    \caption{Compare feature-based KD with 10\% data and 90 epochs. Acc: averaged top-1 accuracy over 4 classification tasks.}
    \vspace{-0.5cm}
    \label{tab: fkd}
 \end{table}

 \subsection{Example: Breaking Relative Magnitude.} \label{sec: eg}
In Sec. \textcolor{red}{3.3} of our paper, we argue that Scale Normalization (SN) and Std Matching (SM) are local transformations that transform channel-wisely, which may break the original relative magnitude between channels. Here, we provide a toy example as an illustration.

For instance, we have a three channel penultimate layer with target Std=[4, 3, 2] and after SVD Std=[50, 5, 1]. For a feature after SVD that is [10, 2, 2], with SN we have [0.2, 0.4, 2], and with SM we have [0.8, 1.2, 4], both losing the original relative magnitude.


\section{Implementation Details} \label{sec: imple}
We implement our method using the PyTorch \cite{paszke2019pytorch} framework and use SGD with momentum of 0.9 for all our experiments. All experiments are conducted using four 32G V100 GPUs.

\subsection{KDEP Setups}
For the KDEP procedure, we use an initial learning rate (lr) of 0.3 for R50$\rightarrow$R18 and 0.1 for R50$\rightarrow$MNv2. Batch size is set to 512.  For data augmentation, we use RandomResizedCrop(224) and RandomHorizontalFlip.
In order to reduce the burden of hyper-parameter tuning (\eg weight decay), we multiply our feature-based loss (refer to Eq. \textcolor{red}{1} in our paper) by a loss weight $w$, which matches the feature-based loss to the loss scale of supervised pretraining. For reproduction, we provide the loss weight of different teacher-student pairs in Table \ref{taba: loss weight}. 

\begin{table}[htbp] 
    \centering
    \vspace{-0.0cm}
    \begin{small}
        \setlength\tabcolsep{6pt} 
        \begin{tabular}{l|l}
            \hline
            Teacher$\rightarrow$Student & $w$    \\ 
            \hline
            Standard SP R50$\rightarrow$R18 & 20 \\
            MS R50$\rightarrow$R18 & 3 \\
            SWSL R50$\rightarrow$R18 & 1 \\
            MEAL V2 R50$\rightarrow$R18 & 1 \\
            Swin-B$\rightarrow$R18 & 3 \\
            MS R50$\rightarrow$MNV2 & 3 \\
           \hline
        \end{tabular}
    \end{small}
    \caption{Value of loss weight $w$ for different T-S pairs.}
    \vspace{-0.3cm}
    \label{taba: loss weight}
\end{table}

 For the 10\% ImageNet data setting, 
 we sample 10\% images from each class of the original 1000 class in ImageNet-1K.
 We set the training epochs to 90 or 180, and drop the lr by a factor of 10 at 1/3 and 2/3 of total epochs. When using all of ImageNet data, we train for 9 or 18 epochs to verify fast convergence, and for 90 epochs to further boost performance, where 90 epochs with all ImageNet data is the standard supervised pretraining schedule \cite{salman2020adversarially}. We use weight decay $\in{\{\text{1$e$-4, 4$e$-4, 5$e$-4}\}}$ according to the length of the training schedule, shown in Table \ref{taba: wd}. 
 
 \begin{table}[htbp] 
    \centering
    \vspace{-0.0cm}
    \begin{small}
        \setlength\tabcolsep{6pt} 
        \begin{tabular}{cc|c}
            \hline
            Data & Epoch & Weight decay    \\ 
            \hline
            10\% & 90 & 5$e$-4 \\
            10\% & 180 & 4$e$-4 \\
            100\% & 9 & 5$e$-4 \\
            100\% & 18 & 4$e$-4 \\
            100\% & 90 & 1$e$-4 \\
            \hline
        \end{tabular}
    \end{small}
    \caption{Weight decay for different KDEP training scedules.}
    \vspace{-0.3cm}
    \label{taba: wd}
\end{table}

\subsection{Downstream Task Setups}
For all downstream tasks, we use the same schedule and evaluation protocols for all models for a fair comparison.

For image classification, we initialize the backbone with the distilled weights and add a linear classifier with random initialization. We train the network for 150 epochs with batch size of 64, weight decay of 5$e$-4, an initial learning rate $\in{\{\text{0.01, 0.001}\}}$ which drops by a factor of 10 at 1/3 and 2/3 of total epochs. Again, we use RandomResizedCrop(224) and RandomHorizontalFlip for data augmentation.

For semantic segmentation, we also initialize the backbone with the distilled weights, and add a PSP module \cite{zhao2017pyramid} and a segmentation head after the backbone. We use batch size of 16, an initial learning rate of 0.01, weight decay of 1$e$-4, crop size of 512, and deploy an polynomial learning rate annealing procedure \cite{chen2017deeplab}. For data augmentation, we use random scaling, random horizontal flipping, random rotation, and random Gaussian blur.
The number of epochs is 50, 100, 200 for VOC12, ADE20K, and Cityscapes, respectively, following previous standard~\cite{zhao2017pyramid}.

For object detection, we experiment with the Faster R-CNN \cite{ren2015faster} detector and backbones are also initialized by the distilled weights. Unless noted, all the setups follow the evaluation protocols in MOCO \cite{he2020momentum}. For ResNet18, we use a backbone of R18-C4 (similar to R50-C4 \cite{he2017mask}) for both VOC and COCO experiments. For MobileNetV2, we equip it with a FPN \cite{lin2017feature} backbone. 1x schedule is applied for COCO. 

\subsection{Parametric/Non-parametric Methods.}
In our experiments, we experiment with three 1$\times$1 conv variants for parametric methods: Post-ReLU, Pre-ReLU and the one in \cite{heo2019comprehensive}. 
Specifically, we add a 1$\times$1 convolutional layer and a Batch Normalization layer either after (Post-ReLU) or before (Pre-ReLU) the ReLU activation function. We also experiment with the Pre-ReLU 1$\times$1 conv method equipped with Margin ReLU of teacher's feature and Partial $L_2$ loss function as in \cite{heo2019comprehensive}.  

For non-parametric methods, we explore channel selection (CS.var, CS.rand), interpolation, and SVD. For channel selection methods, we experiment with two methods: selecting the top-$D_s$ channels with largest variances (CS.var) or random selecting $D_s$ channels (CS.rand). For interpolation method, we use the default nearest-neighbor interpolation in PyTorch \cite{paszke2019pytorch}. For SVD, we calculate the singular vectors offline and use the top-$D_s$ principal components to transform the teacher's features during the online KDEP process.

\end{document}